\documentclass{article}
\usepackage{spconf,amsmath,graphicx}
\usepackage{subfigure}
\usepackage[utf8]{inputenc}
\usepackage{overpic}
\usepackage{xcolor}
\usepackage{makecell}
\usepackage{booktabs}
\usepackage{tikz}
\usepackage{amssymb}
\usepackage{verbatim}
\usepackage{enumitem}
\usepackage[hyphens]{url}
\usepackage{hyperref}
\usepackage[outline]{contour}
\usepackage{tcolorbox}
\usepackage{soul}

\usepackage{footmisc}

\title{Toward Sufficient Spatial-Frequency Interaction for Gradient-aware UNDERWATER IMAGE ENHANCEMENT}

\name{Chen Zhao, Weiling Cai$^{\dag}$, Chenyu Dong, Ziqi Zeng}
\vspace{-9pt}
\vspace{-9pt}
\address{School of Artificial Intelligence, Nanjing Normal University, Nanjing, 210023,  China}

\vspace{-9pt}

\begin{document}
%\ninept
\small

\maketitle
%\footnotetext[1]{This work was supported by the National Natural Science Foundation of China (Grant No. 62276138).}
\begin{abstract}
Underwater images suffer from complex and diverse degradation, which inevitably affects the performance of underwater visual tasks. However, most existing learning-based underwater image enhancement (UIE) methods mainly restore such degradations in the spatial domain, and rarely pay attention to the fourier frequency information. In this paper, we develop a novel UIE framework based on spatial-frequency interaction and gradient maps, namely SFGNet, which consists of two stages. Specifically, in the first stage, we propose a dense spatial-frequency fusion network (DSFFNet), mainly including our designed dense fourier fusion block and dense spatial fusion block, achieving sufficient spatial-frequency interaction by cross connections between these two blocks. In the second stage, we propose a gradient-aware corrector (GAC) to further enhance perceptual details and geometric structures of images by gradient map.  Experimental results on two real-world underwater image datasets show that our approach can successfully enhance underwater images, and achieves competitive performance in visual quality improvement. The code  is available at \href{https://github.com/zhihefang/SFGNet}{https://github.com/zhihefang/SFGNet}.

\end{abstract}
%\mycopyrightnotice

\vspace{-6pt}
\section{Introduction}
\vspace{-6pt}
Acquiring high-quality underwater images is crucial for underwater tasks, such as underwater robotics and ocean resource exploration. However, underwater images suffer from degradation due to the absorption and scattering of light \cite{AkkaynakTSLTI17}, and this inevitably affects the performance of underwater visual tasks. Underwater image enhancement (UIE) aims to obtain high-quality images by removing scattering and correcting color distortion in degraded images \cite{ag2017underwater}.
To address this problem, traditional UIE methods based on the physical properties of the underwater images were proposed  \cite{peng2017underwater,drews2013transmission,yan2018}. These methods investigate the physical mechanism of the degradation caused by color cast or scattering and compensate them to enhance the underwater images. 
However, these physics-based model with limited representation capacity cannot address all the complex physical and optical factors underlying the underwater scenes, which leads to poor enhancement results under highly complex and diverse underwater scenes. \footnotetext[1]{This work was supported by the National Natural Science Foundation of China (Grant No. 62276138).}
Recently, some learning based methods for UIE can produce better results, since neural networks have powerful feature representation and nonlinear mapping capabilities. It can learn the mapping of an image from degenerate to clear from a substantial quantity of paired training data. Yang et al. \cite{jian2018} proposed a reflected light-aware multi-scale progressive restoration network to obtain images with both color equalization and rich texture in various underwater scenes. A U-Net with spatial- and channel-wise normalization was employed to handle the variability of underwater scenes \cite{zhenqi2022}. Ma et al. \cite{ziyin22} further presented a wavelet-based dual-stream network that addresses color cast and blurry details. However, most of those methods are based on the spatial information and rarely pay attention to the Fourier frequency information, which has been proved to be effective for low-light image restoration \cite{chun23}.

Inspired by previous Fourier-based works \cite{chun23}, we explore the properties of the Fourier frequency information for UIE task. Fig. 1 shows our motivation. Given two images (underwater image and its corresponding normal-light label image), we swap their amplitude components and combine them with corresponding phase components in the Fourier space. The recombined results show that the visual appearance are swapped following the amplitude swapping, which indicates the degradation information of underwater images is mainly contained in the amplitude component. Consequently, underwater images can be enhanced by amplitude component in the Fourier space. Moreover, the Fourier transform does not introduce massive parameters, which is very efficient. %\footnotetext[]{This work was supported by the National Natural Science Foundation of China (Grant No. 62276138).}

\begin{figure}[t]
	\centering
	\includegraphics[width=1.0\linewidth]{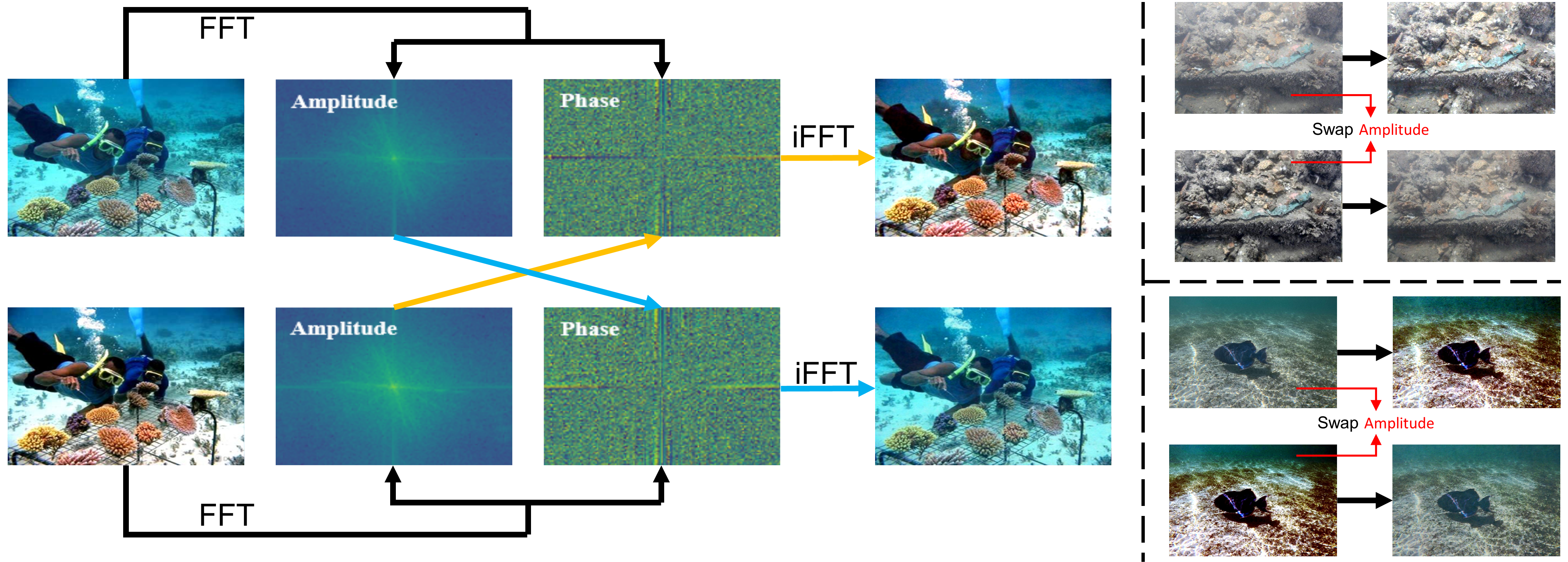}
	\vspace{-9pt}
	\vspace{-9pt}

	\caption{
		Our motivations. The amplitude and
		phase are produced by Fast Fourier Transform (FFT) and the compositional images are obtained by Inverse FFT
		(IFFT). We swap the amplitude components of the underwater image and its corresponding label image. The recombined result of the amplitude of label image
		and the phase of underwater image has similar visual appearance with label image.
	}
	\label{fig:framework}
	\vspace{-9pt}
\end{figure}

\begin{figure*}[t]
	\includegraphics[width=\linewidth]{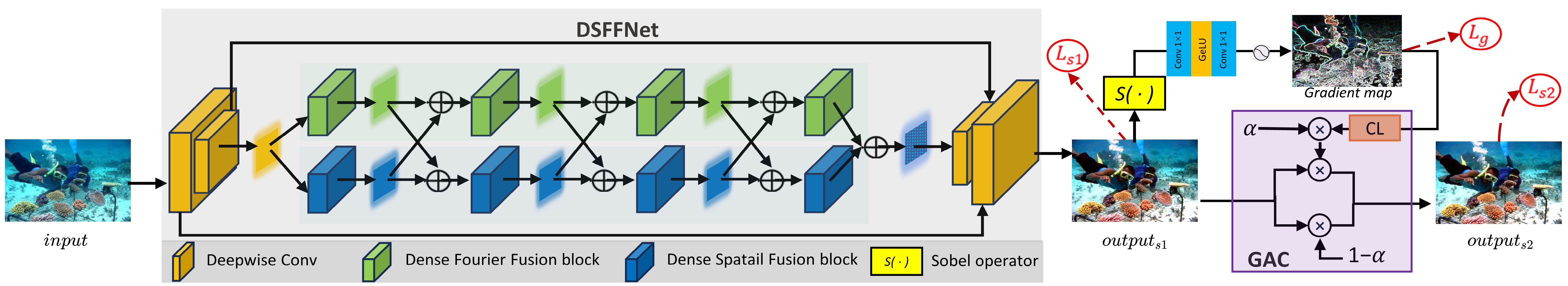}
	\vspace{-9pt}
	\vspace{-9pt}

	\label{fig:network}
	\caption{Overall framework of our proposed SFGNet.  SFGNet mainly consists of a dense spatial-frequency fusion network (DSFFNet) and gradient-aware corrector (GAC). In GAC, CL refers to the curricular learning strategy. DSFFNet aims to achieve sufficient spatial-frequency interaction by cross connections between dense fourier fusion (DFF) block and dense spatial fusion (DSF) block. To further refine the details, we propose a novel gradient-aware corrector (GAC) to further enhance perceptual details and geometric structures of images by gradient map.}
	\vspace{-9pt}
	\vspace{-3pt}

\end{figure*}

In this paper, we develop a novel UIE framework by incorporating spatial-frequency interaction and gradient maps, called SFGNet, which mainly consists of two stages.
In the first stage, a dense spatial-frequency fusion network (DSFFNet) is proposed. Specifically, DSFFNet mainly includes our designed dense fourier fusion block and dense spatial fusion block, achieving sufficient spatial-frequency interaction by cross connections between these two blocks. In the stage, our goal is to achieve enhancement of amplitude component. Gradient maps can not only help deep networks pay attention to geometric structures \cite{cheng2017underwater}, but also  capture edges and textures of images. Consequently, we propose a gradient-aware corrector to further enhance perceptual details of images by gradient map in the second stage. In summary, the main contributions are as follows:

\begin{itemize}[leftmargin=2em]
	\setlength{\parskip}{-1pt}
	\item
	We propose a novel UIE framework based on the fourier transform and gradient maps, %spatial-frequency and gradient domain,
	 namely SFGNet, which can generate high-quality underwater images. 
	
	\item We present a DSFFNet that can achieve spatial-frequency interaction to enhance amplitude
	component, which takes full advantage of the property of the fourier frequency information. 
	
	\item
	A gradient-aware corrector (GAC) is proposed, which further enhances edges and textures of images, and enriches the details and structure of the images.
	
\end{itemize}

\begin{figure}[t]
	\centering
	\includegraphics[width=1.0\linewidth]{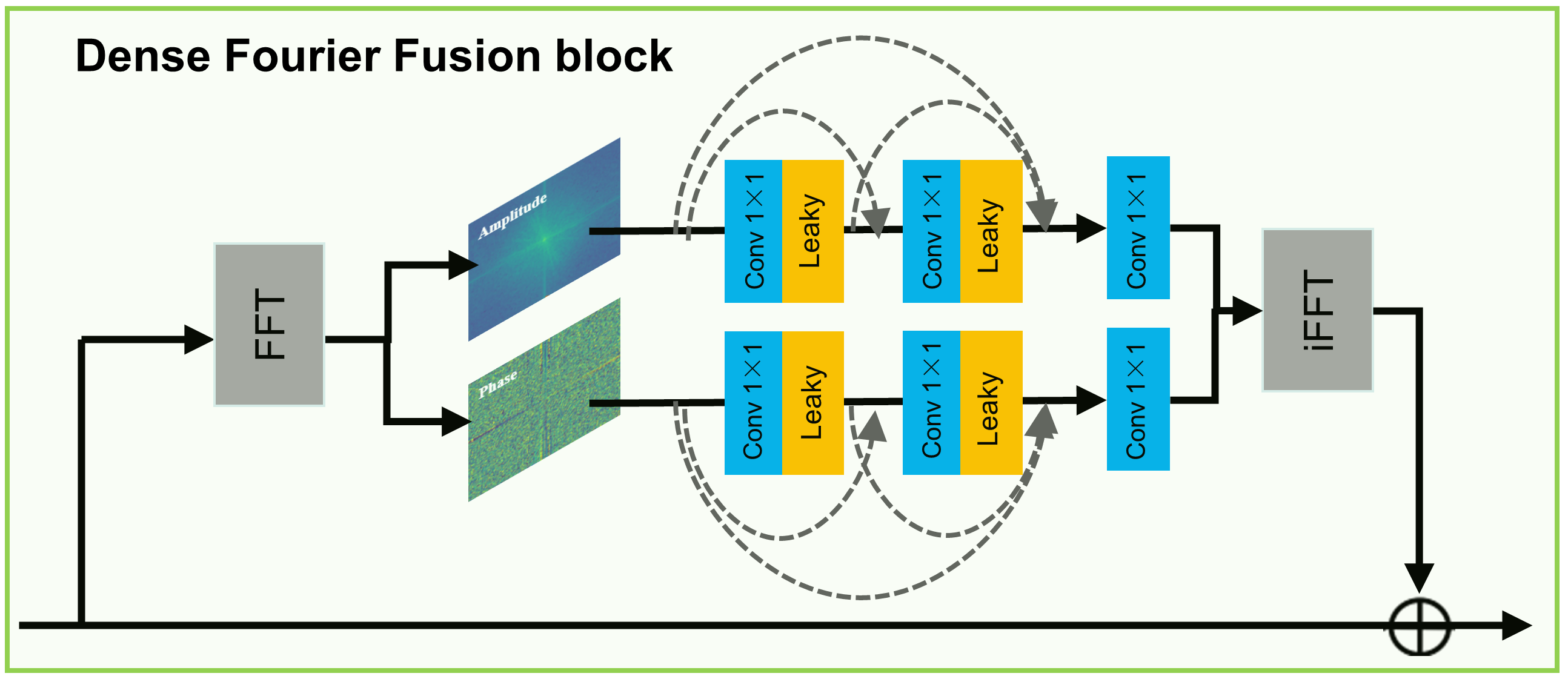}
	\vspace{-9pt}
	\vspace{-9pt}

	\caption{
		The detailed structure of the designed DFF block. 
	}
	\label{fig:framework}
	\vspace{-3pt}
	\vspace{-9pt}
	
\end{figure}

\begin{figure}[t]
	\centering
	\includegraphics[width=1.0\linewidth]{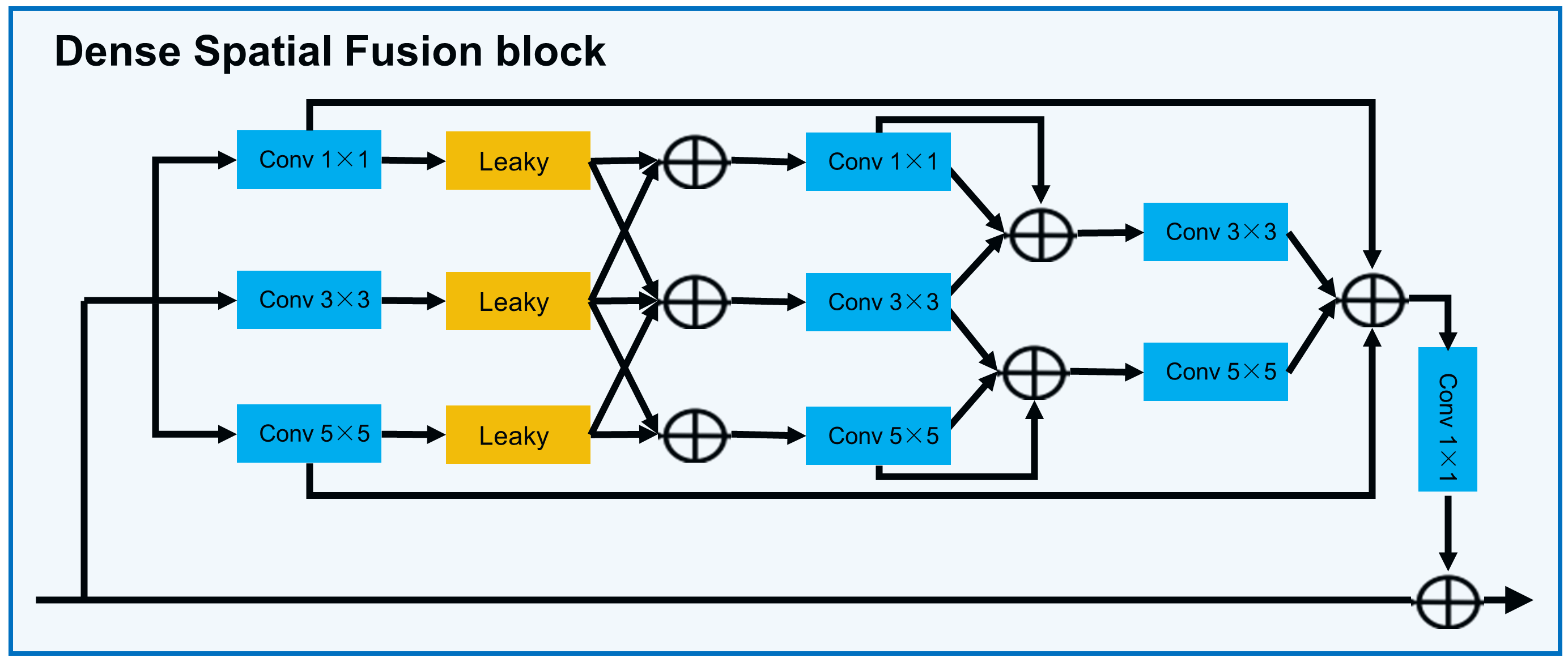}
	
	\vspace{-9pt}
	
	\caption{
		The detailed structure of the designed DSF block. 
	}
	\label{fig:framework}
	\vspace{-9pt}

\end{figure}

\vspace{-9pt}
\section{Proposed method}
\vspace{-3pt}
\subsection{ Overall Framework}
Given an underwater image as input, we aim to learn a network to generate an enhancement output that removes the color cast from input while enhancing the image details. The framework of SFGNet is shown in Fig. 2, which mainly consists of a dense spatial-frequency fusion network (DSFFNet) and gradient-aware corrector (GAC). DSFFNet mainly consists of our designed dense fourier fusion (DFF) block and dense spatial fusion (DSF) block, achieving sufficient spatial-frequency interaction by cross connections between these two blocks. We can obtain the $output_{s1} $ via DSFFNet, and contrain the network using $\mathcal{L}_{s1}$. We mainly employ the Sobel operator to obtain the gradient map, and adaptively obtain the $output_{s2} $ via GAC. Note that, we adopt a curricular learning strategy to make the GAC converge better.

\subsection{ Fourier Transform}
Firstly, we briefly introduce the operation of the Fourier transform \cite{g2017underwater}.  Given a single channel image $x$, whose shape is $H\times W$, the Fourier transform $\mathcal{F}$ which converts $x$ to the Fourier space $X$ can be expressed as:
\begin{equation}
\scriptsize
\mathcal{F}\left(x\right)(u,v)=X(u,v)=
\frac{1}{\sqrt{HW}}\sum_{h=0}^{H-1}\sum_{w=0}^{W-1}x(h,w)e^{-j2\pi(\frac{h}{H}u+\frac{w}{W}v}),
\end{equation}
where $h, w$ are the coordinates in the spatial space and $u, v$ are the
coordinates in the Fourier space. $\mathcal{F}^{-1}$ denotes the inverse transform of $\mathcal{F}$. 

Complex component $X(u,v)$ in the Fourier space can be represented
by a amplitude component $\mathcal{A}(X(u,v))$ and a phase component $\mathcal{P}(X(u,v))$ as follows:
\begin{equation}
\begin{array}{l}\mathcal{A}(X(u,v))=\sqrt{R^2(X(u,v))+I^2(X(u,v))},\\ \mathcal{P}(X(u,v))=arctan[\frac{I(X(u,v))}{R(X(u,v))}],\end{array}
\end{equation}
where the $R(X(u, v))$ and $I(X(u, v))$ can be obtained by:
\begin{equation}
\begin{matrix}R(X(u,v))=\mathcal{A}(X(u,v))\times cos(\mathcal{P}(X(u,v)))\\ I(X(u,v))=\mathcal{A}(X(u,v))\times sin(\mathcal{P}(X(u,v)))\end{matrix}
\end{equation}
where $R(x)$ and $I(x)$ represent the real and imaginary parts of $X(u, v)$. $X(u, v)$ can also be represented by:
\begin{equation}
X(u,v)=R(X(u,v))+jI(X(u,v))
\end{equation}

According to our motivation and Fig.1, we conclude that the degradation information of underwater images is mainly contained in the amplitude component. Consequently, in spatial-frequency interaction stage, we improve the underwater images by constraining the amplitude component in the Fourier space. To further refine the detail information, we propose a gradient-aware corrector to further enhance perceptual details of images by gradient map.
%Wavelet transform has been applied to various image processing tasks such as image super-resolution~\cite{huang2017wavelet} and denoising~\cite{kang2018deep}.
%Several physics-based methods in underwater image enhancement have used DWT to decompose the images and process them in the frequency domain~\cite{singh2014enhancement, priyadharsini2018wavelet} to improve the contrast and resolution. In our framework, we use DWT to decompose an input image into multiple frequency sub-bands so that the color correction and detail enhancement can be separately addressed.
\subsection{ Dense Spatial-Frequency Fusion Network}

Based on the above analysis, we design a simple but effective DSFFNet with an encoder-decoder format. To enhance the performance of the network and reduce the parameters of the network, we mainly use  depthwise separable convolution. In order to obtain rich frequency and spatial information, we designed dense fourier fusion (DFF) block and dense spatial fusion (DSF) block, which are shown Fig. 3 and Fig. 4. At the bottleneck, we employ four DFF blocks and four DSF blocks to extract hidden frequency and spatial information. In order to achieve interactive information exchange between the spatial and frequency domains, we use cross connections between these two blocks to further extract spatial-frequency fusion information.

Let us denote $x_{in}$ and $x_{out1}$ as the $input$ and $output_{s1}$ of the DSFFNet, $x_{gt}$ represents the ground truth, and their representations in Fourier space are denoted as $X_{in}$, $X_{out1}$ and $X_{gt}$, respectively. To learn the amplitude representation in the Fourier space, we constrain the amplitude component by supervised manner. The loss function for the spatial-frequency interaction stage $L_{s1}$ is expressed as:
\begin{equation}
\mathcal{L}_{s1}=||\mathcal{A}(X_{out1})-\mathcal{A}(X_{gt})||_1,
\end{equation}
where $\mathcal{A}()$ refers to the amplitude component in Fourier transform.
\vspace{-5pt}
\subsection{ Gradient-Aware Corrector}
To further refine the details, we propose a novel gradient-aware corrector (GAC) to further enhance perceptual details and geometric structures of images by gradient map. Gradient maps can not only help deep networks pay attention to geometric structures, but also  capture edges and textures of images. Firstly, we employ the Sobel operator to calculate the gradient map of $output_{s1}$. After that, we apply a two-layer convolution mapping network to enhance the edge details of the gradient map. The gradient map $G_{s1}$ of $output_{s1}$ ($x_{out1}$), which can be expressed as:
\begin{equation}
G_{s1}=Sigmoid(Conv(GeLU(Conv(S(x_{out1})))))
\end{equation}
where $S(.)$ represent the Sobel operator. Note that, to improve the edge details of the gradient map, we constrain the gradient map by supervised manner. Let us denote $G_{gt}$ represents the gradient map of ground truth. The loss function for the gradient map $L_{g}$ is expressed as:
\begin{equation}
	\mathcal{L}_g=\|G_{s1}-G_{gt}\|_2
\end{equation}

In order to make the model converge better, we adopt a curricular learning strategy \cite{yong2017underwater} to train the network. During training, the $G_{s1}$ and $G_{gt}$ are fused to generate the final gradient map $G$. $G$ can be expressed mathematically as:
\begin{equation}
G=\lambda\cdot G_{s1}+(1-\lambda)\cdot G_{gt},
\end{equation}
where $\lambda$ is the trade-off parameter, and can be dynamically
adjusted through $\mathcal{L}_g$:
\begin{equation}
\lambda=\begin{cases}0,&\text{if } \mathcal{L}_g>0.1,\\ \frac{\mathcal{L}_g-0.1}{0.1-0.05},&\text{if } 0.1\ge\mathcal{L}_g>0.05,\\ 1,&\text{if } \mathcal{L}_g\le0.05.\end{cases}
\end{equation}

After obtaining the gradient map $G$, we adaptively
weight the $output_{s1}$ ($x_{out1}$) through a learnable parameter $\alpha$. The $output_{s2}$ ($x_{out2}$) can be obtained by
\begin{equation}
x_{out2}=(1-\alpha)\otimes x_{out1}+\alpha\cdot G\otimes x_{out1},
\end{equation}
where $\otimes$ being the operator of pixel-wise multiplication.
We set the loss function $L_{s2}$  for the gradient-aware corrector:
\begin{equation}
\mathcal{L}_{s2}=||x_{out2}-x_{gt}||_{1}+||\phi(x_{out2})-\phi(x_{gt})||_{1},
\end{equation}
where $\phi$ is a pre-trained VGG network \cite{ka2018deep}.
Finally, the overall loss $\mathcal{L}_{total}$ of SFGnet can be expressed as :
\begin{equation}
\mathcal L_{total}=\gamma_1\mathcal L_{s2}+\gamma_2\mathcal L_{s1}+\gamma_3\mathcal{L}_g
\end{equation}
where $\gamma_1$,$\gamma_2$, and $\gamma_3$ are the hyper-parameters. The
best performance is achieved when we set $\gamma_1$,$\gamma_2$, and $\gamma_3$ to 1, 0.1 and 0.5.
\vspace{-9pt}
\section{EXPERIMENTS}
\vspace{-3pt}

\begin{table}[t]

\caption{Quantitative comparison on the UIEBD datasets and LSUI datasets. The best results are boldfaced. $\uparrow$: The higher, the better, $\downarrow$: the lower, the better. }
\centering

\resizebox{0.95\linewidth}{!}{
\begin{tabular}{l| c c c| c c c}
\specialrule{1.2pt}{0.2pt}{1pt}
\multicolumn{1}{c}{Methods} & \multicolumn{3}{c}{UIEBD~\cite{guo2017underwater}} & \multicolumn{2}{c}{LSUI\cite{lin2017underwater}}\\
\midrule
\multicolumn{1}{c}{Name}   & PSNR $\uparrow$ & SSIM $\uparrow$ & LPIPS $\downarrow$ & PSNR $\uparrow$ & SSIM $\uparrow$\\
\midrule

UIECˆ2-Net~\cite{yudong2017underwater} & 20.14 & 0.8215 & 0.2033 & 20.86 &  0.8867\\
Water-Net~\cite{guo2017underwater}  & 19.35 & 0.8321 & 0.2116 & 19.73 & 0.8226\\
UIEWD~\cite{ziyin22}  & 14.65 & 0.7265  & 0.3956 & 15.43 & 0.7802  \\
UWCNN~\cite{saeed2018deep}  & 15.40 & 0.7749 & 0.3525 & 18.24 & 0.8465\\
SCNet~\cite{zhenqi2022}  & 20.41 & 0.8235 & 0.2497 & 22.63 & 0.9176\\
U-shape~\cite{lin2017underwater}  &21.25 & 0.8453 & 0.1977 & 22.86 & 0.9142\\
\midrule
Ours  & \textbf{21.66 }& \textbf{0.8710} & \textbf{0.1909} & \textbf{23.56} & \textbf{0.9361}\\
\specialrule{1.2pt}{0.2pt}{1pt}
\end{tabular}}
\label{tab:quan}

\end{table}

\begin{figure*}[t]
	\includegraphics[width=\linewidth]{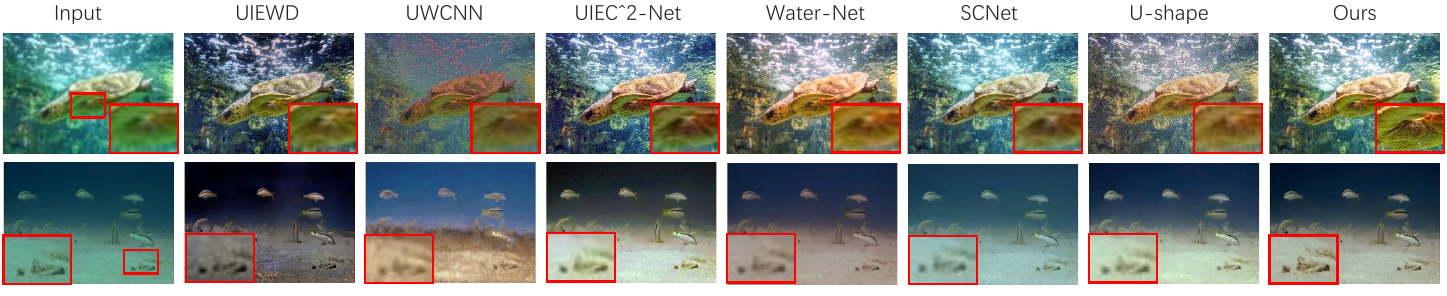}
	\vspace{-9pt}
	\vspace{-9pt}
	\vspace{-5pt}

	\label{fig:network}
	\caption{ Comparisons of visual results with all baselines on the UIEBD dataset. Red box marked image is a zoomed image. %The first row and second row are two examples on the UIEBD dataset, and the third row and fourth row are two examples on the LSUI dataset.
	}
\vspace{-9pt}
\vspace{-3pt}

\end{figure*}

\begin{figure*}[t]
	\includegraphics[width=\linewidth]{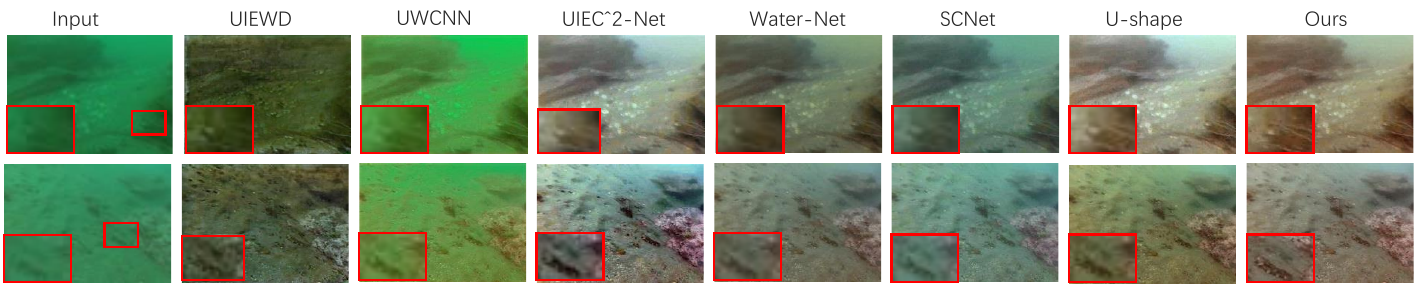}

	\vspace{-9pt}
	\vspace{-9pt}

	\label{fig:network}
	\caption{ Comparisons of visual results with all baselines on the LSUI dataset. Red box marked image is a zoomed image %The first row and second row are two examples on the UIEBD dataset, and the third row and fourth row are two examples on the LSUI dataset.
	}

\end{figure*}

\subsection{Setup}

% \noindent
% \textbf{Implementation details.}
Our network, implemented with PyTorch 1.7, was trained and tested on an NVIDIA Geforce RTX 3090 GPU. We use the Adam optimizer, $\beta_{1}= 0.9$ and $\beta_{2}= 0.999$. Our patch size was set to 256 ×
256 and batch size to 5. We set the initial learning rate to 0.001, which is steadily decreased to 0.000001 using the cosine annealing strategy \cite{chilov2018deep}. We employ the real-world UIEBD dataset \cite{guo2017underwater} and  the LSUI dataset \cite{lin2017underwater} to train and test our model. UIEBD contains 890 underwater images
and corresponding labels. We use the 700 images for training and the rest for testing. The LSUI dataset was randomly divided as 4500 images for training and 400 images for testing. We compare SFGNet
with six SOTA UIE methods including UIECˆ2-Net \cite{yudong2017underwater}, Water-Net \cite{guo2017underwater} and UIEWD \cite{ziyin22}, UWCNN \cite{saeed2018deep}, SCNet \cite{zhenqi2022} and U-shape \cite{lin2017underwater}. We mainly adopt the well-known full-reference image quality assessment metrics: PSNR and SSIM \cite{zw2017underwater}. PSNR and SSIM quantitatively compare
our method with other methods in
terms of pixel level and structure level, respectively. The higher the
values of the PSNR, SSIM, the better the quality of the generated
images. In addition, we also use the LPIPS metric for evaluation. LPIPS \cite{chardTSLTI17} is a deep neural network-based image quality metric.  It compares the human visual system's perception of an image with a reference image and provides a score that reflects the degree of visual similarity. A lower LPIPS score means a better UIE approach.

\subsection{Results and comparisons}

Table 1 shows the quantitative results compared with different baselines
on UIEBD dataset and LSUI dataset, including with UIECˆ2-Net, Water-Net, UIEWD, UWCNN, SCNet, U-shape. We mainly use PSNR, SSIM and LPIPS as our quantitative indices for UIEBD dataset, and employ PSNR and SSIM as our quantitative indices for LSUI dataset. The results in Table 1 show that our algorithms outperform state-of-the art methods obviously, and achieve state-of-the-art performance in terms of three full-reference
image quality evaluation metrics on
UIE task, which verifies the robustness of the proposed SFGNet. To better validate the superiority of our methods, in
Fig. 5 and Fig. 6, we show the visual results comparison with
state-of-the-art methods on  UIEBD dataset and LSUI dataset, respectively. The two examples in Fig. 5 are two cases randomly selected on the UIEBD dataset, and the two examples in Fig. 6 are two cases randomly selected on the LSUI dataset.
Our methods consistently generate natural and better
visual results on testing images,
strongly proving that  SFGNet has good
generalization performance for real-world applications.

\begin{table}[t]
\caption{Ablation study without GAC on UIEBD dataset. }
\centering

\resizebox{0.85\linewidth}{!}{
\begin{tabular}{c c c  c c}
\specialrule{1.2pt}{0.2pt}{1pt}
\multicolumn{3}{c}{Methods} &  \multicolumn{2}{c}{UIEBD~} \\
\cmidrule(lr){1-3}
\cmidrule(lr){4-5}
DFF block & DSF block & SFI  & PSNR $\uparrow$ & SSIM$\uparrow$ \\
\midrule
\midrule
\checkmark & &   & 19.76  & 0.8213\\
 & \checkmark& & 19.21 &0.8175\\
\checkmark & \checkmark &  & 20.07& 0.8389\\
\checkmark & \checkmark & \checkmark & \textbf{20.97} & \textbf{0.8425}\\
\specialrule{1.2pt}{0.2pt}{1pt}
\end{tabular}}
\label{tab:loss}

\end{table}

\begin{table}
	\caption{Ablation study with GAC on UIEBD dataset.}
	\centering
	\label{tab:freq}

	\begin{tabular}{c|cccc|cc}
		\specialrule{1.2pt}{0.2pt}{1pt}
		
		Method  &$\mathcal{L}_{s1}$ & CL  & $ \mathcal{L}_{g} $  & $\alpha$ & PSNR$\uparrow$ &SSIM $\uparrow$         \\\midrule
		
		A&$\times$           &  $\checkmark$             & $\checkmark$         &$\checkmark$         & 20.13 &  0.8365      \\
		B&$\checkmark$  & $\times$                    & $\checkmark$         & $\checkmark$         &21.45 & 0.8641       \\

		C&$\checkmark$          &  $\times$                & $\times$        & $\checkmark$         & 21.12& 0.8573       \\
        D&$\checkmark$          &  $\checkmark$                & $\checkmark$       & $\times$         & 20.81& 0.8426       \\
		\midrule

		Full model &$\checkmark$   & $\checkmark$        &  $\checkmark$             &  $\checkmark$         &   \textbf{21.66}     & \textbf{0.8710}      \\
		
		\specialrule{1.2pt}{0.2pt}{1pt}
		
	\end{tabular}

\end{table}

\subsection{Ablation study}
\textbf{Ablation study without GAC.}
In order to evaluate the effectiveness of our proposed DFF block and DSF block, we conduct an ablation study with GAC on the UIEBD dataset in Table 2. SFI is spatial-frequency interaction by cross connections between these two blocks. Note that, we do not discuss the effect of our proposed GAC here. We regularly remove one component to each configuration at one time, and our strategy achieves the best performance by using all blocks and SFI, proving that sufficient spatial-frequency interaction is useful for UIE task.

\noindent
\textbf{Ablation study with GAC.}
In this section, we will discuss the effectiveness of all loss functions and strategies, and Table 3 shows the quantitative results on
 the UIEBD dataset. CL refers to the curricular learning strategy the final gradient map $G$. $\alpha$ represents the adaptively
  learnable parameter of the $output_{s2}$. The results of model A indicate that the constraint $\mathcal{L}_{s1}$ by supervised learning manner on the amplitude component is effective. Compared with the full model, model B obtains lower indicators, proving that the curricular learning strategy is useful for the gradient maps. The results of model C mean that the gradient map can be improved by supervised learning strategy. Model D not achieves satisfactory results, indicating that the gradient information can be adaptively obtained via the learnable parameter $\alpha$.

\vspace{-9pt}
\section{Conclusion}
\vspace{-9pt}
\label{sec:concl}

In this paper, we develop a novel UIE framework based on spatial-frequency interaction and gradient maps, namely SFGNet, which  accurately removes the degradation information of the underwater images. SFGNet consists of a dense spatial-frequency fusion network (DSFFNet) and gradient-aware corrector (GAC). DSFFNet mainly aims to %including our designed dense fourier fusion block and dense spatial fusion block, 
achieve spatial-frequency interaction by cross connections between DFF  and DSF blocks. GAC is proposed to further enhance perceptual details and geometric structures of images by gradient maps. Equipped with the above techniques, our algorithm can show SOTA performance on two real-world underwater image datasets, and achieves competitive performance in visual quality. %In fact, our proposed framework is a general scheme based on the fourier transform and gradient information in low-level vision tasks. Consequently, in the future we will explore the potential the fourier transform and gradient information in other tasks.

\end{document}